\documentclass[runningheads]{llncs}
\usepackage[T1]{fontenc}
\usepackage{graphicx}
\usepackage[shortlabels]{enumitem}
\usepackage{xcolor}
\usepackage{color}
\definecolor{mygray}{RGB}{193,203,215}
\usepackage{booktabs}
\usepackage{multirow}
\usepackage{colortbl}
\usepackage{amsmath}
\usepackage{subcaption}
\usepackage{graphicx}
\begin{document}
\title{LoRA-Enhanced Vision Transformer for Single Image based Morphing Attack Detection via
Knowledge Distillation from EfficientNet}
%
%
\author{Ria Shekhawat\inst{1}\orcidID{0009-0006-9558-4337} \and Sushrut Patwardhan\inst{1} \and Raghavendra Ramachandra\inst{1}\orcidID{0000-0003-0484-3956} \and
 Praveen Kumar Chandaliya \inst{2}\orcidID{0000-0002-4787-0308} \and Kishor P. Upla \inst{2}\orcidID{0000-0001-6306-0682}
}
\authorrunning{Ria et al.}
%
\institute{Norwegian University of Science and Technology (NTNU), Norway. \\ \and
 Sardar Vallabhbhai National Institute of Technology, Surat, India. 
\\
\email{Email: raghavendra.ramachandra@ntnu.no}\\
}
\maketitle              
\begin{abstract}
Face Recognition Systems (FRS) are critical for security but remain vulnerable to morphing attacks, where synthetic images blend biometric features from multiple individuals. We propose a novel Single-Image Morphing Attack Detection (S-MAD) approach using a teacher-student framework, where a CNN-based teacher model refines a ViT-based student model. To improve efficiency, we integrate Low-Rank Adaptation (LoRA) for fine-tuning, reducing computational costs while maintaining high detection accuracy. Extensive experiments are conducted on a morphing dataset built from three publicly available face datasets, incorporating ten different morphing generation algorithms to assess robustness. The proposed method is benchmarked against six state-of-the-art S-MAD techniques, demonstrating superior detection performance and computational efficiency.

\end{abstract}
\section{Introduction}
In the era of automation, Face Recognition Systems (FRS) have become an integral part of security and authentication, from mobile phone unlocking to passport issuance and Automatic Border Control (ABC). With advances in morphing techniques, FRS becomes increasingly vulnerable to morphing attacks, as highlighted in \cite{8053499},\cite{venkatesh2021face}. Morphing attacks manipulate FRS by presenting a synthetically generated image that combines biometric samples from two or more images, allowing multiple individuals to gain unauthorized access.


Growing concerns over morphing attacks have led to the development of Morphing Attack Detection (MAD) techniques, broadly categorized into Single-image MAD (S-MAD) and Differential MAD (D-MAD). Detecting attacks from a single image (S-MAD) is more challenging than comparing images (D-MAD), making S-MAD a critical focus for robust security. Recent advancements in S-MAD leverage foundation models, identity disentanglement, generative modeling, transformers, and multi-contributor analysis to enhance detection accuracy. Caldeira et al. \cite{caldeira2025madationfacemorphingattack} fine-tuned CLIP with Low-Rank Adaptation (LoRA) to improve feature alignment for morph detection. Caldeira et al. \cite{caldeira2023unveilingtwofacedtruthdisentangling} introduced an autoencoder-based identity disentanglement approach, enabling classifiers to distinguish morphed identities. Ivanovska et al. \cite{ivanovska2023mad_ddpm} employed Denoising Diffusion Probabilistic Models (DDPMs) to model bona fide distributions, detecting morphed images as out-of-distribution samples. Zhang et al. \cite{zhang2024ViT} utilized Vision Transformers (ViTs) to enhance feature extraction for cross-dataset morph detection. Tapia et al. \cite{tapia2023alphanet} introduced AlphaNet, which detects morphing attacks from multiple contributors, addressing variations in morph complexity. Sushrut et al. \cite{MADCLIP} presented a multimodal learning approach that can provide a textual description of morphing attack detection using CLIP. Despite these advancements, generalization remains a challenge across diverse morphing techniques, as highlighted by Zhang et al. \cite{zhang2024ViT}. Strengthening S-MAD methods is crucial to ensuring robust and adaptable morphing attack detection in real-world applications.

Enhancing morph detection while reducing computational costs can be achieved through Knowledge Distillation (KD), where, conventionally, a powerful teacher model transfers knowledge to a smaller, efficient student model. In this work, we propose a novel S-MAD technique based on the teacher-student framework, focusing on selecting effective pre-trained teacher and student networks. Given the demonstrated success of Vision Transformers (ViTs) in face morph detection \cite{zhang2024ViT}, our approach utilizes a CNN-based teacher model (EfficientNetV2) to refine a ViT-based student model, improving detection accuracy while maintaining computational efficiency. Our setup inverts the traditional KD paradigm, which involves a large teacher model distilling knowledge to a smaller student. This decision is motivated by several factors: First, ViTs have recently shown superior performance in fine-grained tasks like face morphing detection due to their global attention mechanism, making them highly suitable as high-capacity learners in this domain. Second, EfficientNetV2, while computationally lightweight, benefits from strong inductive biases and stable generalization capabilities learned from large-scale datasets, making it an effective and consistent knowledge provider. The teacher doesn’t necessarily have to be bigger, instead, it needs to be stable and well-generalized for the task. Additionally, by integrating LoRA (Low-Rank Adaptation), we enable efficient fine-tuning of the ViT student by updating only a small subset of parameters, reducing computational overhead while maintaining high detection performance. Thus, our method prioritizes task-specific effectiveness and training efficiency, enabling robust knowledge transfer from CNN to Transformer and ensuring practicality in deployment. As a result, the proposed method is expected to be both computationally efficient and highly reliable in detecting morphing attacks. The main contributions of this paper are:
\begin{itemize} [leftmargin=*,noitemsep, topsep=0pt,parsep=0pt,partopsep=0pt]
\item A novel S-MAD technique is proposed, based on a teacher-student architecture with LoRA optimization, ensuring computational efficiency and robust morphing attack detection.
\item Extensive experiments are conducted on three morphing datasets, created using publicly available face datasets such as FRGC \cite{Phillips-OverviewFaceRecognitionGrandChallengeFRGC-CVPR-2005}, FERET \cite{phillips1998feret} and FRLL \cite{DeBruine2017FaceRL}. These datasets include ten different morphing techniques, ranging from landmark-based to generative AI-based methods.
\item The proposed method's quantitative performance is evaluated and compared against six recently published state-of-the-art methods.
\end{itemize}
The rest of the paper is organized as follows: Section \ref{sec:pro} presents the proposed S-MAD algorithm, Section \ref{sec:Exp} discusses the morphing databases, evaluation protocol, implementation details, and quantitative results and Section \ref{sec:Concl} draws the conclusion.

\section{Proposed Method}
\label{sec:pro}

In this section, we introduce a novel approach for Single-Image Morphing Attack Detection (S-MAD) leveraging Knowledge Distillation (KD) in a student-teacher framework. We used EfficientNetV2 as the teacher model due to its proven efficiency and high performance in feature extraction, making it ideal for training on large-scale datasets with various morphing attack types. The architecture of EfficientNetV2, optimized for both accuracy and computational efficiency, provides robust embeddings that guide the student model effectively.

For the student model, we opt for the Vision Transformer (ViT) fine-tuned with Low-Rank Adaptation (LoRA), capitalizing on ViT’s ability to capture long-range dependencies in image data. The integration of LoRA improves the efficiency of the fine-tuning process, allowing the student to learn from the teacher's rigor while maintaining computational scalability. This combination of EfficientNetV2 and ViT with LoRA ensures that the student model benefits from the teacher’s rich feature representations while optimizing for better generalization and performance in morphing attack detection.

The key innovation of the proposed method lies in fine-tuning the student model using teacher embeddings via an adapter, allowing it to outperform the teacher model in detection accuracy.
This approach highlights the effectiveness of knowledge transfer in enhancing S-MAD performance, improving both detection accuracy and adaptability. Furthermore, the proposed method demonstrates strong generalization across a wide range of morphing generation techniques, highlighting its robustness and establishing a promising foundation for future advancements in S-MAD. 
\begin{figure}[htp]
	\centering
	\includegraphics[width = 1\linewidth]{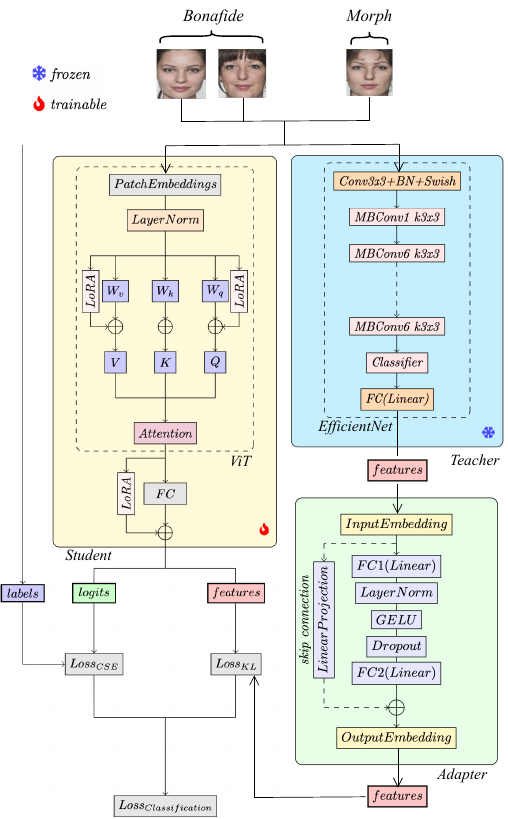}
	\caption{\textbf{Overview of proposed method}: Bonafide and Morph images are passed through the frozen pre-trained teacher and active trainable student models. The feature embeddings from the teacher are passed to the adapter, which transforms and passes them for knowledge distillation to the KL loss function that also receives the student's features. Logits from the student and true labels are used to compute the CE loss. These two losses make the combined loss function for the student.}
    \label{fig:figure1}
\end{figure}

The block diagram of the proposed method is illustrated in Figure \ref{fig:figure1}. The approach utilizes KD in a teacher-student framework, where learned features from a frozen, pre-trained teacher model are processed through an adapter to transfer essential knowledge to the student model. An efficient teacher model enables the student model to train more effectively, reducing computational costs while enhancing detection accuracy. This method presents a promising direction for leveraging larger student models in S-MAD, optimizing both performance and efficiency.
At the core of this methodology is the integration of Cross-Entropy (CE) loss with Kullback-Leibler (KL) divergence, which minimizes the difference between the projected embeddings of the student and teacher models. By reducing the divergence between the softened teacher and student features, this approach enhances the model’s ability to distinguish bona fide faces from morphed ones. This combination effectively guides the student model in learning class-specific representations that closely align with the teacher’s embeddings, improving generalization across different morphing attack types and strengthening robustness against novel attacks. Additionally, the adapted embeddings, combined with LoRA’s efficient fine-tuning, contribute to a more effective morphing attack detection model, achieving higher accuracy with reduced computational complexity. In the following, we discuss the teacher, student and adapter architectures of the proposed method. 

\textbf{Teacher Network:}
To train the student model, we first obtain feature embeddings from the teacher model, which is a pre-trained convolution network (EfficientNetV2), fine-tuned on a large S-MAD dataset with various morph types. Details on the training dataset for teacher are discussed in Section \ref{sec:Exp}.

\textbf{Adapter Network:}
The adapter module processes teacher embeddings, transforming them to effectively distill and align features within the student model’s embedding space. The adapter applies a series of transformations, including fully connected layers, normalization, activation functions, and dropout, ensuring that critical features are retained while adapting to the student's representation. A residual connection is incorporated to add the original input embedding back to the output, facilitating better gradient flow and preserving essential information while ensuring alignment with the student model’s feature space. This approach improves training stability and efficiency, enabling the student model to retain relevant knowledge from the Teacher embeddings while effectively leveraging the transformation for enhanced morphing attack detection.

\textbf{Student Network:}
The student model is a pre-trained Vision Transformer (ViT), fine-tuned using Low-Rank Adaptation (LoRA) to efficiently adapt to the task of morphing attack detection. It leverages embeddings generated by the teacher model, which has been trained on a separate dataset, facilitating cross-domain knowledge transfer.
During fine-tuning, the student model learns to align its embeddings with the teacher’s representations while simultaneously optimizing its classification logits. The loss function is designed to incorporate both embedding similarity between the teacher and student models and the student’s classification logits, ensuring an effective knowledge transfer mechanism via the adapter module. This approach enhances the student model's ability to generalize across diverse morphing attack types.

\subsection{Training Procedure and Loss Function}
The training process consists of two key stages: fine-tuning the teacher model and fine-tuning the student model, using separate datasets for each phase.
In the first stage, the teacher model is fine-tuned on DB1 (described in Section \ref{sec:Exp}). In the second stage, during student model training, the fine-tuned teacher generates embeddings for input face images from DB2, which are then processed through the adapter before being passed to the loss function.
By leveraging KL divergence, the adapted embeddings help the student model learn a discriminative feature space, ensuring that embeddings of bona fide faces remain closer, while those of morphing attacks are pushed farther apart. This results in a fine-tuned student model that outperforms the teacher in detecting morphing attacks.

The proposed Knowledge Distillation (KD) framework minimizes a unified loss function that jointly optimizes knowledge transfer and student model classification performance. The loss function comprises two key components: (1) Kullback-Leibler (KL) divergence loss, which aligns the teacher’s and student’s embeddings, ensuring effective knowledge transfer, and (2) Cross-Entropy (CE) loss, which optimizes the student model’s classification accuracy.
By balancing these terms, the framework enhances the student model’s ability to learn discriminative features while improving its morphing attack detection performance.

Let the student logits be denoted as \( \mathbf{z}_s \) and the ground truth labels as \( y \). The student's classification loss is computed using the standard cross-entropy loss:
\begin{equation}
\vspace{-2mm}
L_{\text{CE}}(\mathbf{z}_s, y) = -\sum_{i} y_i \log(\hat{y}_i)
\end{equation}
where \( \hat{y}_i \) is the predicted probability for class \( i \) obtained by applying softmax to \( \mathbf{z}_s \).

For the distillation process, the teacher's embeddings \( \phi_t(f_t) \) are passed through an adapter network to obtain the adapted teacher embeddings \( \phi_a(f_t) \). The student embeddings are denoted by \( \phi_s(f_s) \). To measure the alignment between the student and teacher outputs, we compute the Kullback-Leibler (KL) divergence between their softened probability distributions.

To soften the logits and obtain probability distributions, both the adapted teacher and student embeddings are passed through a softmax function with a temperature parameter \( T \):
\begin{equation}
\vspace{-2mm}
\tilde{p}_t = \text{softmax} \left( \frac{\phi_a(f_t)}{T} \right), \quad \tilde{p}_s = \text{softmax} \left( \frac{\phi_s(f_s)}{T} \right)
\end{equation}
where \( \tilde{p}_t \) and \( \tilde{p}_s \) represent the softened class probability distributions for the teacher and student, respectively.

The KL loss is then given by:
\begin{equation}
\vspace{-2mm}
L_{\text{KL}}(\tilde{p}_t, \tilde{p}_s) = \sum_{j=1}^{C} \tilde{p}_{t,j} \log \left( \frac{\tilde{p}_{t,j}}{\tilde{p}_{s,j}} \right)
\end{equation}
where \( C \) is the number of classes, and \( \tilde{p}_{t,j} \), \( \tilde{p}_{s,j} \) are the softened probabilities for class \( j \) from the teacher and student, respectively.

The KL loss then computes the divergence between these softened probability distributions. It is weighted by a hyperparameter \( \lambda \) (set to 0.5), which controls the relative contribution of the KL loss. The final classification loss is then the weighted sum of the KL loss and the cross-entropy loss:
\vspace{-2mm}
\begin{equation}
\vspace{-2mm}
\text{loss} = \lambda \cdot L_{\text{KL}}(\tilde{p}_t, \tilde{p}_s) + L_{\text{CE}}(\mathbf{z}_s, y)
\end{equation}


By minimizing this loss function, the student model is trained to both accurately classify the data and mimic the teacher's embeddings after adaptation.

\section{Experimental Results and discussion}
\label{sec:Exp}
This section presents a quantitative evaluation of the proposed method in comparison with existing Single-Image Morphing Attack Detection (S-MAD) techniques, providing a detailed performance analysis. The detection metrics follows the ISO/IEC SC 37 30107 standard \cite{ISO-IEC-30107-3-PAD-metrics-170227}, utilizing key performance metrics to measure effectiveness. The Morphing Attack Classification Error Rate (MACER) evaluates the percentage of morphing attacks incorrectly classified as genuine images, while the Bona Fide Presentation Classification Error Rate (BPCER) quantifies the proportion of authentic images mistakenly identified as morphing attacks. Additionally, the Detection-Equal Error Rate (D-EER) is reported, which signifies the point where MACER and BPCER are equal, providing a balanced measure of the detection system’s overall reliability.
\subsection{Morphing Datasets}
\begin{figure}[htp]
	\centering
	\includegraphics[width = 1\linewidth]{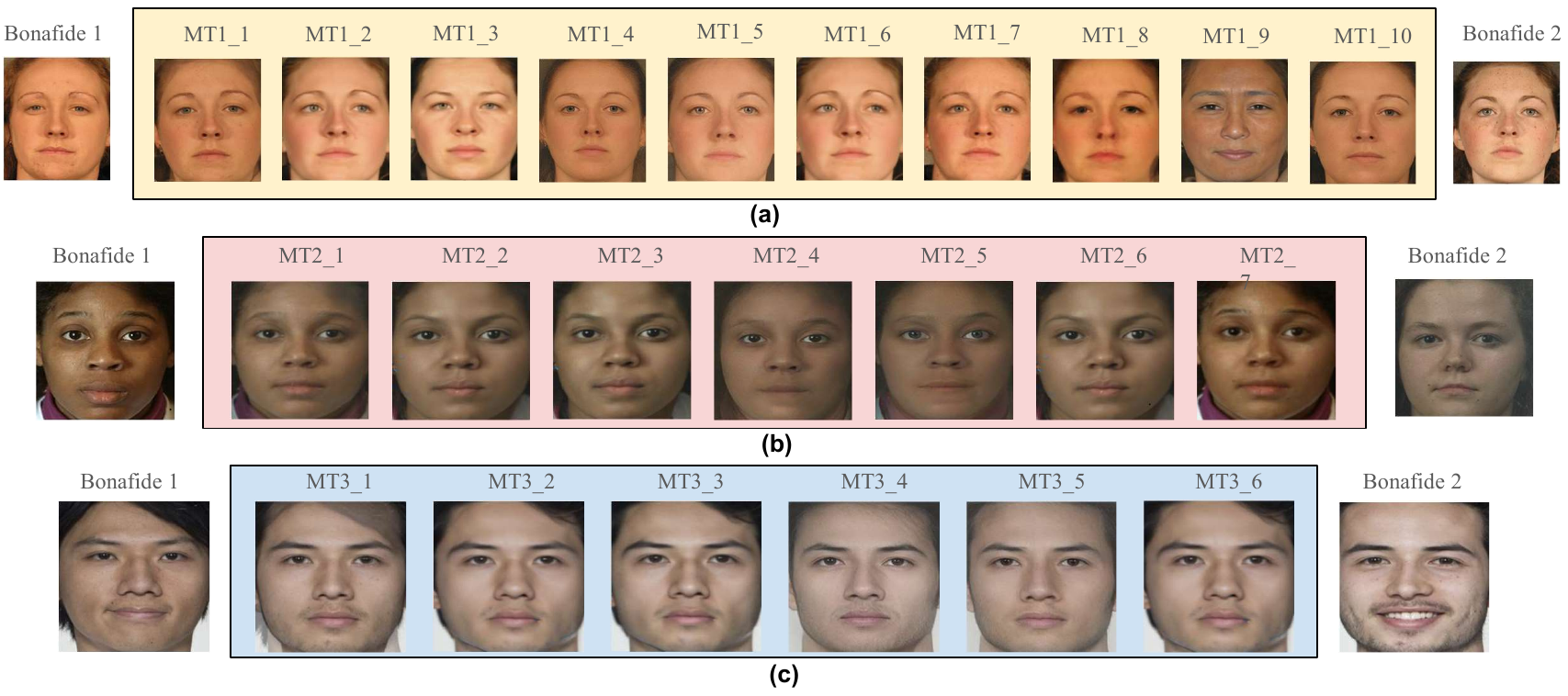}
	\caption{Illustration of bona fide and morphing images from DB1, DB2 and DB3 with different types of Morphing generation Techniques (MT).}
	\label{fig:db}
\end{figure}


In this work, we have employed three publicly available face datasets such as FRGC \cite{Phillips-OverviewFaceRecognitionGrandChallengeFRGC-CVPR-2005} (DB1), FERET \cite{phillips1998feret} (DB2) and FRLL \cite{DeBruine2017FaceRL} (DB3) to generate the morphing images. DB1 morphing dataset is constructed using 147 unique data subjects and ten different morphing techniques are employed to generate the morphing images.  The morphing generation techniques include: Landmark based morphing generation \cite{Landmark-face-morph} ($MT1\_1$), Diffusion Model with identity loss based morphing generation (PIPE) \cite{PIPE} ($MT1\_2$), Diffusion model with greedy approach \cite{blasingame_greedy_dim} ($MT1\_3$), MIPGAN-1 based morphing generation \cite{zhang-MIPGAN-TBIOM-2021} ($MT1\_4$), MIPGAN-2 based morphing generation \cite{zhang-MIPGAN-TBIOM-2021} ($MT1\_5$), Diffusion Model based morphing generation  (MorDiff)\cite{MoDiff} ($MT1\_6$), Landmark based morphing generation with post-processing \cite{Ferrara-TextureBlendingAndShapeWarpingInFaceMorphing-IEEE-BIOSIG-2019} ($MT1\_7$), Visual Code based morphing generation (MorCode) \cite{pn2024morcode} ($MT1\_8$), Extended Latent Mapper \cite{CVMIMorphGeneration} ($MT1\_9$) and StyleGAN-2 based morphing generation \cite{venkatesh-StyleGAN-morph-IWBF-2020}  ($MT1\_{10}$). For DB2, we have used the first 7 morph generation techniques described above for DB1, whereas, for DB3, we have used the first 6.


 DB1 has a total of 1276 bona fide images and $1197 \times 10 = 11970$ morphing samples, while DB2 has 904 unique data subjects resulting in 2003 bona fide and $4449 \time 7 = 31143$ morphing images and DB3 has a total of 102 bona fide images and $1135 \time 6 = 6810$ morphing images.
\subsection{Performance evaluation protocol}
In this work, we have used DB1 to train the teacher network of the proposed method and the student network is fine-tuned using DB2, while DB3 is used only to evaluate the detection performance. Similar protocol is also used with state-of-the art for comparison. 

\subsection{Implementation details}
 To mitigate class imbalance, in the preprocessing pipeline, augmentation is applied more aggressively to the bona fide class, which has fewer samples. Each bona fide image undergoes augmentation twice as often as those in the morph class. Transformations and augmentations are applied to increase dataset variability, improve model generalization, and help counter class imbalance while preserving label integrity.

The Teacher model is based on EfficientNetV2-S, pre-trained on ImageNet-1K and fine-tuned for binary classification on DB1.  The model is trained for 30 epochs with a batch size of 64 using the Adam optimizer with a learning rate of $1 \times 10^{-4}$ and a cosine annealing scheduler with a minimum learning rate of $1 \times 10^{-5}$. The loss function is cross-entropy, and gradient updates utilize mixed precision training with an automatic gradient scaler. Early stopping is implemented with a patience of 5 epochs. Additionally, the model is evaluated every epoch on a validation set from DB1, comprising multiple morphing techniques, and training is halted if validation loss stops improving. 

The Student, ViT-Base (patch16-224) pre-trained on ImageNet-1K, with LoRA applied to its QKV and fully connected layers, was trained using knowledge distillation from the fine-tuned EfficientNetV2 Teacher. The training was conducted with similar hyperparameters as the Teacher but an initial learning rate of $5 \times 10^{-4}$, decayed using a Cosine Annealing LR scheduler with a minimum learning rate of $1 \times 10^{-5}$. The loss function combined cross-entropy and distillation loss from KL loss, weighted by $\alpha$ = 0.5, with a temperature parameter set to 3.0. The LoRA configuration had a rank of 8, $\alpha$ = 16, and a dropout rate of 0.1.

All experiments are conducted on a server equipped with an AMD EPYC 7643 48-Core Processor and running Ubuntu 16.04 LTS. The system is paired with a GRID A100D-20C GPU (20GB), utilizing CUDA 12.2.0 and cuDNN 9.1.0.70 (with cuDNN for CUDA 11), python=3.9 and pyTorch=2.0.1+cu118. 

\subsection{Results and Discussion}
In this section, we present a detailed evaluation of the performance of the proposed student model for morphing attack detection (S-MAD), comparing it to fine-tuned EfficientNet teacher model as well as six state-of-the-art S-MAD algorithms. The results from the various comparative analyses, as shown in Tables \ref{tab:teacher_comparison} and \ref{tab:my-table}, highlight the significant advantages of the proposed method in terms of generalization and accuracy against morphing attacks. Additionally, the evaluation extends to the performance of the algorithm on different morphing generation techniques, with further insights presented in Table \ref{tab:mtypes-table}.

Table~\ref{tab:teacher_comparison} presents a quantitative comparison of the proposed student model against the pre-trained EfficientNet, pre-trained ViT, and the fine-tuned EfficientNet teacher. The proposed method demonstrates a significant improvement with a D-EER of 3.25\%, compared to 50.00\% for pre-trained EfficientNet, 15.34\% for pre-trained ViT, and 24.81\% for the EfficientNet teacher. In terms of BPCER at 5\% MACER, the student model achieves an outstanding 1.86\%, outperforming the EfficientNet teacher (46.44\%) and pre-trained ViT (32.91\%), and EfficientNet (98.59\%) models. Similarly, at 10\% MACER, the student model remains the top performer with a BPCER of 1.86\%, while the pre-trained EfficientNet, ViT, and teacher model exhibit 94.80\%, 21.50\% and 46.44\% BPCER, respectively. These results highlight the effectiveness of the proposed method, emphasizing the role of knowledge distillation and LoRA-based fine-tuning in enhancing morphing attack detection accuracy.
\begin{table}[htp]
\centering
\caption{Quantitative performance of the proposed student against the teacher}
\label{tab:teacher_comparison}
\begin{tabular}{|l|l|ll|}
\hline
\rowcolor[HTML]{EFEFEF} 
\cellcolor[HTML]{EFEFEF} & \cellcolor[HTML]{EFEFEF} & \multicolumn{2}{l|}{\cellcolor[HTML]{EFEFEF}\textbf{BPCER(\%) @ MACER (\%)}} \\ \cline{3-4} 
\rowcolor[HTML]{EFEFEF} 
\multirow{-2}{*}{\cellcolor[HTML]{EFEFEF}\textbf{MAD Algorithm}} &
  \multirow{-2}{*}{\cellcolor[HTML]{EFEFEF}\textbf{D-EER(\%)}} &
  \multicolumn{1}{l|}{\cellcolor[HTML]{EFEFEF}\textbf{5\% \hspace{1cm}} } &
  \textbf{10\%} \\ \hline

EfficientNet                 & 50.00                    & \multicolumn{1}{l|}{98.59}                                  & 94.80          \\ \hline
ViT                 & 15.34                    & \multicolumn{1}{l|}{32.91}                                  & 21.50          \\ \hline
EfficientNet Teacher                 & 24.81                    & \multicolumn{1}{l|}{46.44}                                  & 46.44          \\ \hline
\textbf{Proposed Method- Student} & \textbf{3.25}            & \multicolumn{1}{l|}{\textbf{1.86}}  & \textbf{1.86}  \\ \hline
\end{tabular}%
\end{table}

We evaluate the detection performance of the proposed and six different existing methods for S-MAD that include Xception based approach benchmarked in SynMAD competition \cite{huber2022syn}, AlphaNet from \cite{tapia2023alphanet}, ViT based S-MAD from \cite{zhang2024ViT}, CLIP based approach from \cite{MADCLIP} and \cite{caldeira2025madationfacemorphingattack}, and Diffusion based approach from \cite{ivanovska2023mad_ddpm}. In Table \ref{tab:my-table}, we observe that (a) Ivanovska et al. with D-EER: 33.28\%, Sushrut et al. with D-EER: 29.96\%, Caldeira et al. with D-EER: 20.07\%, Huber et al. with D-EER: 12.74\% and Zhang et al. with D-EER: 12.75\% show moderate performance but still have higher error rates compared to our method. (b) Tapia et al. \cite{tapia2023alphanet} with D-EER: 8.94\% performs better than the other state-of-the-art methods discussed but still struggles with generalization. (c) The Proposed Method achieves the lowest D-EER: 3.25\%, significantly outperforming all existing approaches. This suggests that the teacher-student framework with LoRA fine-tuning provides better generalization and robustness against morphing attacks. 

\begin{table}[htp]
\centering
\caption{Quantitative performance of the proposed and existing S-MAD}
\label{tab:my-table}
\begin{tabular}{|l|l|ll|}
\hline
\rowcolor[HTML]{EFEFEF} 
\cellcolor[HTML]{EFEFEF} & \cellcolor[HTML]{EFEFEF} & \multicolumn{2}{l|}{\cellcolor[HTML]{EFEFEF}\textbf{BPCER(\%) @ MACER (\%)}} \\ \cline{3-4} 
\rowcolor[HTML]{EFEFEF} 
\multirow{-2}{*}{\cellcolor[HTML]{EFEFEF}\textbf{MAD Algorithm}} &
  \multirow{-2}{*}{\cellcolor[HTML]{EFEFEF}\textbf{D-EER(\%)}} &
  \multicolumn{1}{l|}{\cellcolor[HTML]{EFEFEF}\textbf{5\% \hspace{1cm}} } &
  \textbf{10\%} \\ \hline
  Zhang et al. \cite{zhang2024ViT}              & 12.75                    & \multicolumn{1}{l|}{33.33}                                  & 18.63          \\ \hline
  
  Tapia et al. \cite{tapia2023alphanet}                 & 8.94                     & \multicolumn{1}{l|}{19.61}                                  & 8.82           \\ \hline
Huber et al. \cite{huber2022syn}           & 12.74                    & \multicolumn{1}{l|}{22.54}                                  & 13.75          \\ \hline
Sushrut et al. \cite{MADCLIP}                   & 29.96                    & \multicolumn{1}{l|}{75.19}                                  & 63.16          \\ \hline
Ivanovska et al. \cite{ivanovska2023mad_ddpm}                   & 33.28                    & \multicolumn{1}{l|}{74.51}                                  & 65.69          \\ \hline
Caldeira et al. \cite{caldeira2025madationfacemorphingattack}                   & 20.07                    & \multicolumn{1}{l|}{41.18}                                  & 30.29          \\ \hline

\rowcolor[HTML]{ECF4FF} 
\textbf{Proposed Method} & \textbf{3.25}            & \multicolumn{1}{l|}{\cellcolor[HTML]{ECF4FF}\textbf{1.86}}  & \textbf{1.86}  \\ \hline
\end{tabular}%
\end{table}

\begin{figure}[htp]
	\centering
	\includegraphics[width = 0.5\linewidth]{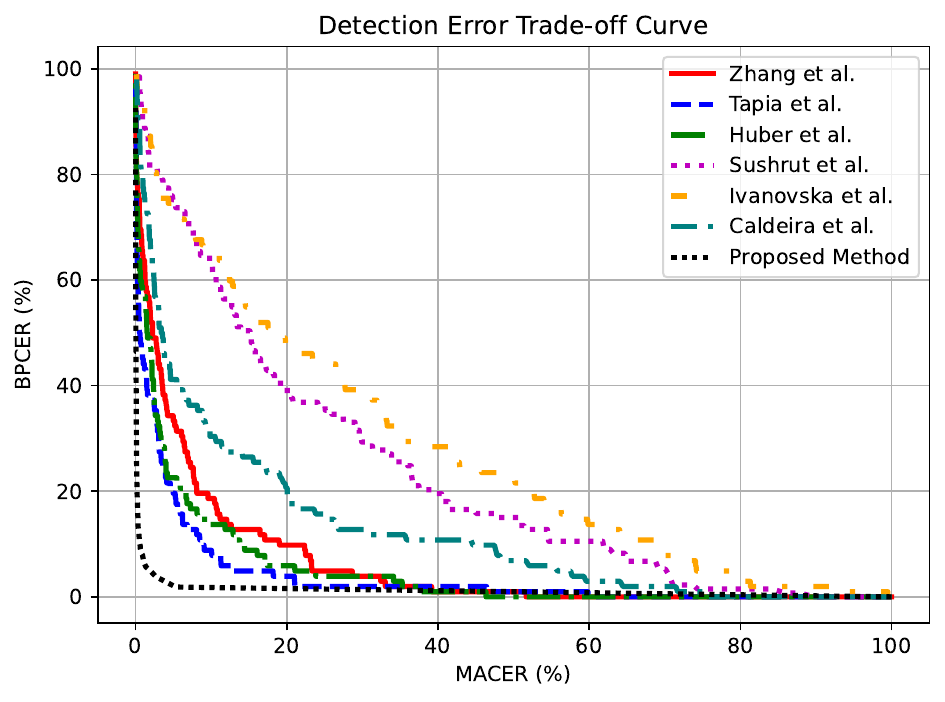}
	\caption{Detection Error Trade-off curve for the Proposed Method against SOTA S-MAD algorithms.}
	\label{fig:det}
\end{figure}


\begin{table}[htp]
\centering
\caption{Quantitative performance of the proposed S-MAD for different Morphing generation Techniques (MT) from DB3}
\label{tab:mtypes-table}
\begin{tabular}{|l|l|ll|}
\hline
\rowcolor[HTML]{EFEFEF} 
\cellcolor[HTML]{EFEFEF} & \cellcolor[HTML]{EFEFEF} & \multicolumn{2}{l|}{\cellcolor[HTML]{EFEFEF}\textbf{BPCER(\%) @ MACER (\%)}} \\ \cline{3-4} 
\rowcolor[HTML]{EFEFEF} 
\multirow{-2}{*}{\cellcolor[HTML]{EFEFEF}\textbf{Morphing Generation Techniques}} &
  \multirow{-2}{*}{\cellcolor[HTML]{EFEFEF}\textbf{D-EER(\%)}} &
  \multicolumn{1}{l|}{\cellcolor[HTML]{EFEFEF}\textbf{5\% \hspace{1cm}} } &
  \textbf{10\%} \\ \hline
  MT3\_1 \cite{Landmark-face-morph}              & 5.59                    & \multicolumn{1}{l|}{7.06}                                  & 1.57          \\ \hline
MT3\_2 \cite{PIPE}                   & 0.65                    & \multicolumn{1}{l|}{0.78}                                  & 0.78          \\ \hline
  MT3\_3
   \cite{blasingame_greedy_dim}                 & 0.95                     & \multicolumn{1}{l|}{1.18}                                  & 1.18           \\ \hline
MT3\_4 \cite{zhang-MIPGAN-TBIOM-2021}           & 4.25                    & \multicolumn{1}{l|}{4.31}                                  & 1.77          \\ \hline
MT3\_5 \cite{zhang-MIPGAN-TBIOM-2021}                   & 3.32                    & \multicolumn{1}{l|}{3.53}                                  & 1.37          \\ \hline
MT3\_6 \cite{MoDiff}                   & 0.88                    & \multicolumn{1}{l|}{1.18}                                  & 1.18          \\ \hline
\end{tabular}%
\end{table}
In Table \ref{tab:mtypes-table}, we evaluate the performance of the proposed S-MAD algorithm across individual morphing generation techniques in DB3. The results demonstrate strong generalization across diverse morph types. Notably, the method performs exceptionally well on diffusion-based morphs MT3\_2, MT3\_3, and MT3\_6 with D-EER values below 1\%, while still maintaining competitive accuracy on more challenging types such as Landmark-based morphs (MT3\_1) and MIPGAN variants MT3\_4, MT3\_5. The consistently low BPCER values at MACER thresholds of 5\% and 10\% reflect the robustness of our approach in detecting morphs generated using different strategies. This underscores the method's applicability in real-world scenarios where morphing techniques vary significantly.
The findings from the comparative analysis clearly demonstrate that the proposed method outperforms existing state-of-the-art approaches in terms of D-EER, BPCER, and MACER, achieving the lowest error rates across various datasets and morphing techniques. These results suggest that the teacher-student framework, combined with LoRA-based fine-tuning, offers superior robustness and generalization capabilities for morphing attack detection. Moreover, the method’s strong performance across a wide range of morphing generation techniques further underscores its applicability in real-world scenarios. Future work may focus on further refining the model's scalability and exploring its integration with other biometric systems for enhanced security.


\section{Explainability and Interpretability}
To enhance the transparency and interpretability of the proposed S-MAD framework, we employ the Local Interpretable Model-Agnostic Explanations (LIME) technique. In the context of face morph detection, explainability plays a crucial role in understanding the decision-making behavior of the model, especially when deployed in high-stakes applications such as biometric authentication.

LIME is particularly well-suited for Vision Transformer (ViT) architectures, as it helps interpret the patch-based input representation and provides intuitive, human-understandable explanations of model predictions. Its localized perturbation-based approach complements the attention mechanisms in ViTs, offering valuable insights into which regions most strongly influence the output.

\begin{figure}[htp]

    \centering
    \fbox{
    \begin{minipage}{0.43\textwidth}
    \begin{subfigure}[t]{0.48\textwidth}
        \centering
        \includegraphics[width=\textwidth]{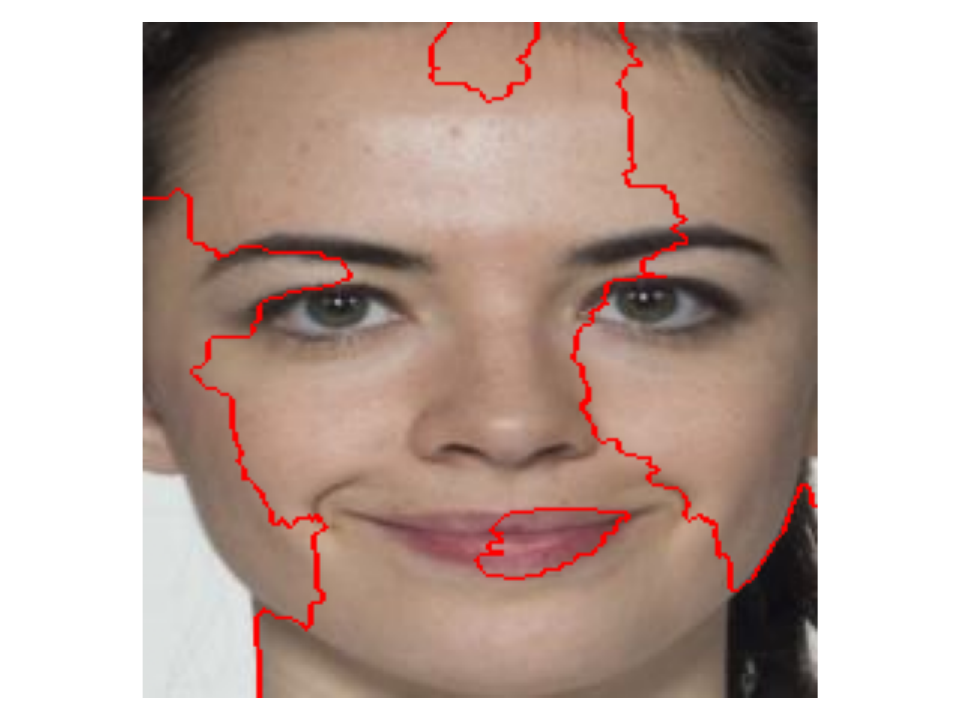}
        \caption{}
        \label{fig:protocol1}
    \end{subfigure} 
    \hspace{0.005\textwidth}
    \begin{subfigure}[t]{0.48\textwidth}
        \centering
        \includegraphics[width=\textwidth]{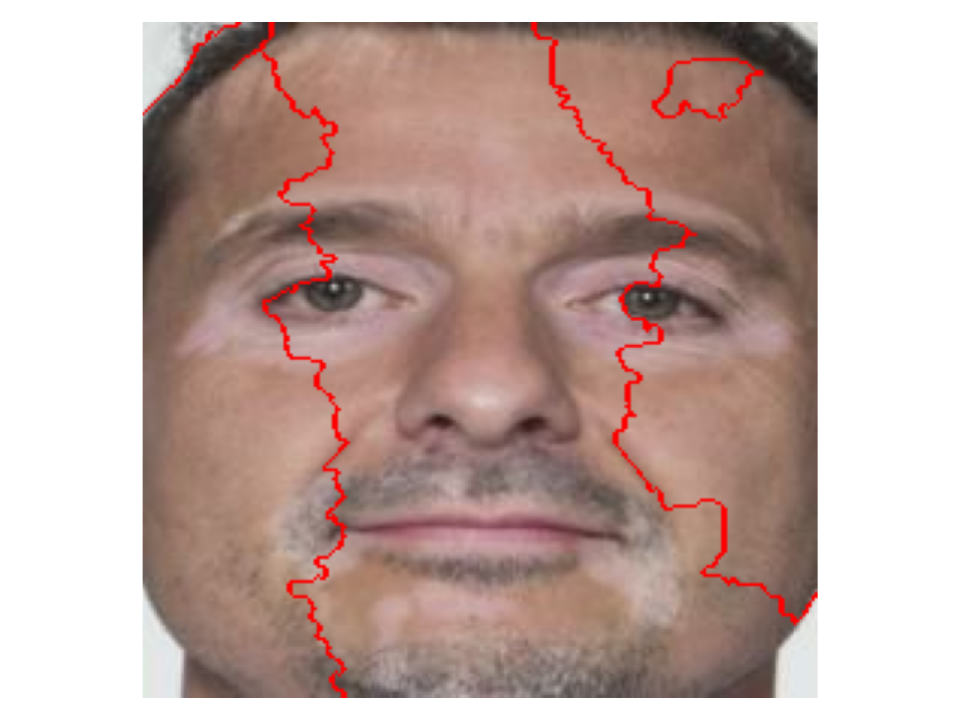}
        \caption{}
        \label{fig:protocol2}
    \end{subfigure}
    \end{minipage}
    }
    \hspace{0.01\textwidth}
    \fbox{
    \begin{minipage}{0.43\textwidth}
    \begin{subfigure}[t]{0.48\textwidth}
        \centering
        \includegraphics[width=\textwidth]{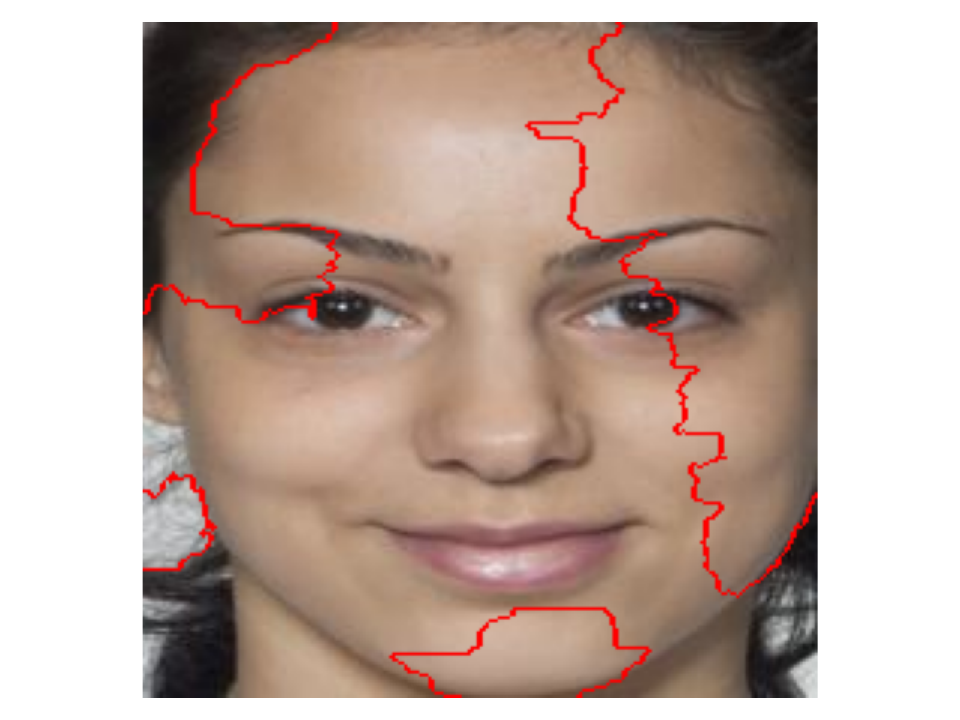}
        \caption{}
        \label{fig:protocol3}
    \end{subfigure}
    \hspace{0.005\textwidth}
    \begin{subfigure}[t]{0.48\textwidth}
        \centering
        \includegraphics[width=\textwidth]{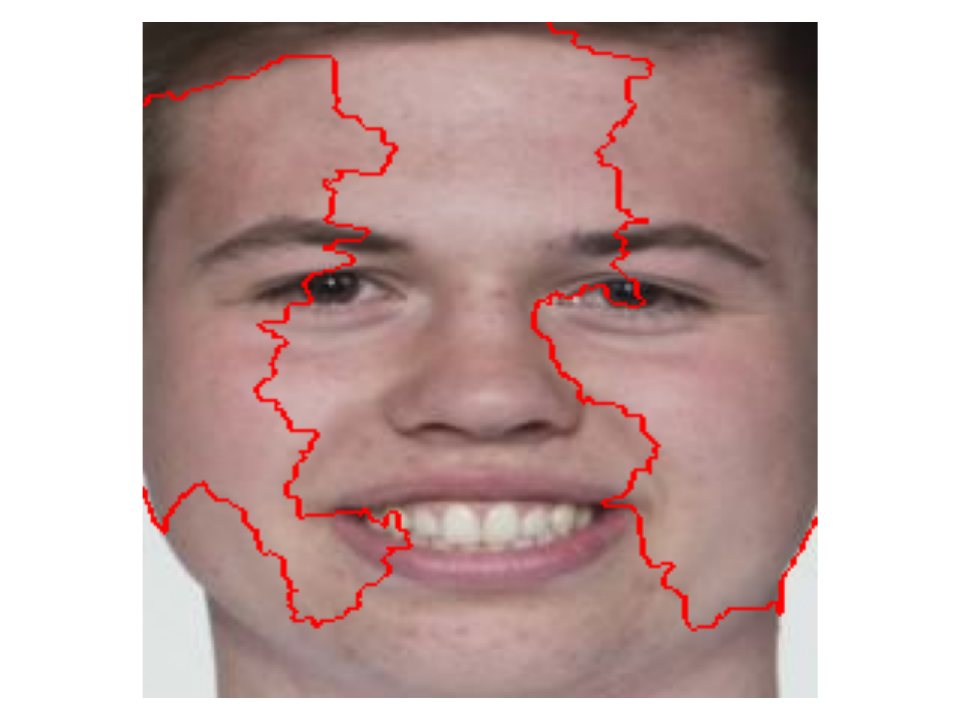}
        \caption{}
        \label{fig:protocol4}
    \end{subfigure}
    \end{minipage}
    }
    \vspace{0.02\textwidth} 

    \fbox{
    \begin{minipage}{0.43\textwidth}
    \begin{subfigure}[t]{0.48\textwidth}
        \centering
        \includegraphics[width=\textwidth]{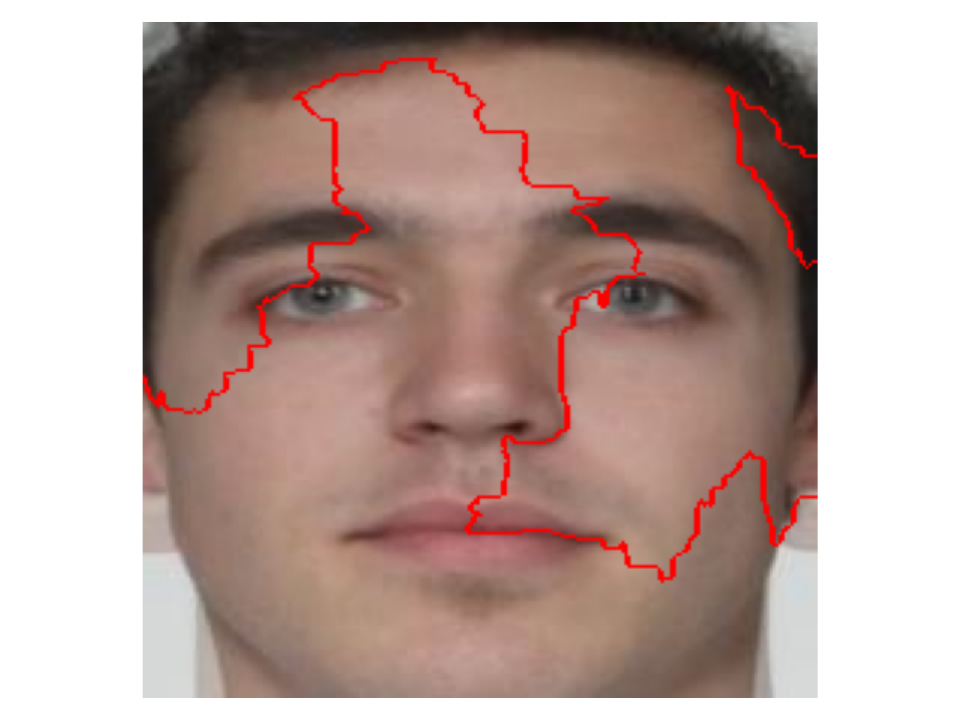}
        \caption{}
        \label{fig:protocol5}
    \end{subfigure} 
    \hspace{0.005\textwidth}
    \begin{subfigure}[t]{0.48\textwidth}
        \centering
        \includegraphics[width=\textwidth]{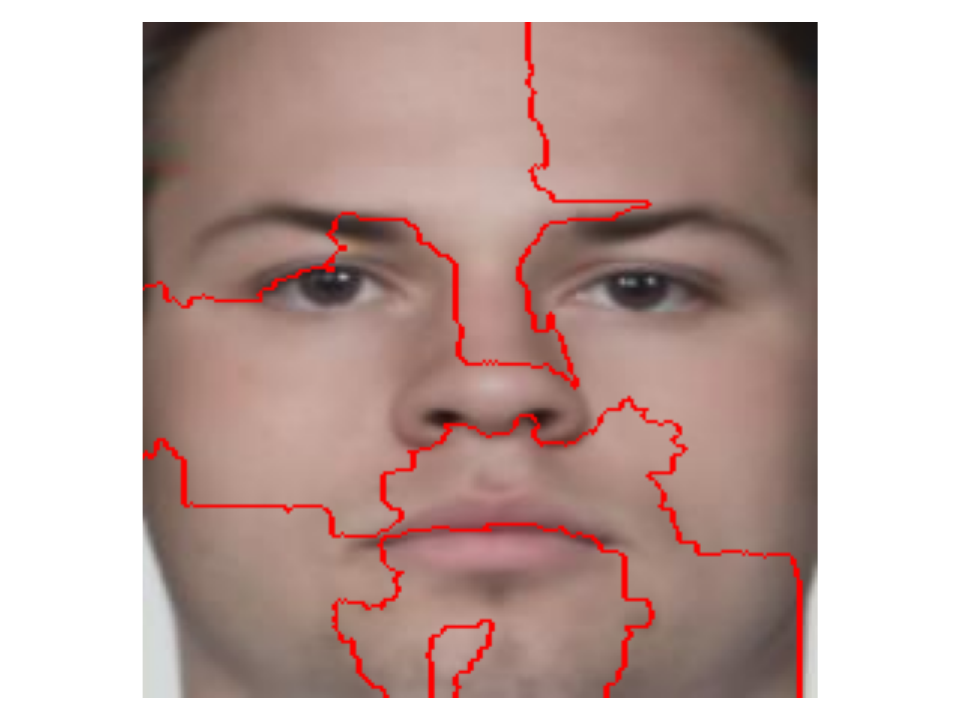}
        \caption{}
        \label{fig:protocol6}
    \end{subfigure}
    \end{minipage}
    }
    \hspace{0.01\textwidth}
    \fbox{
    \begin{minipage}{0.43\textwidth}
    \begin{subfigure}[t]{0.48\textwidth}
        \centering
        \includegraphics[width=\textwidth]{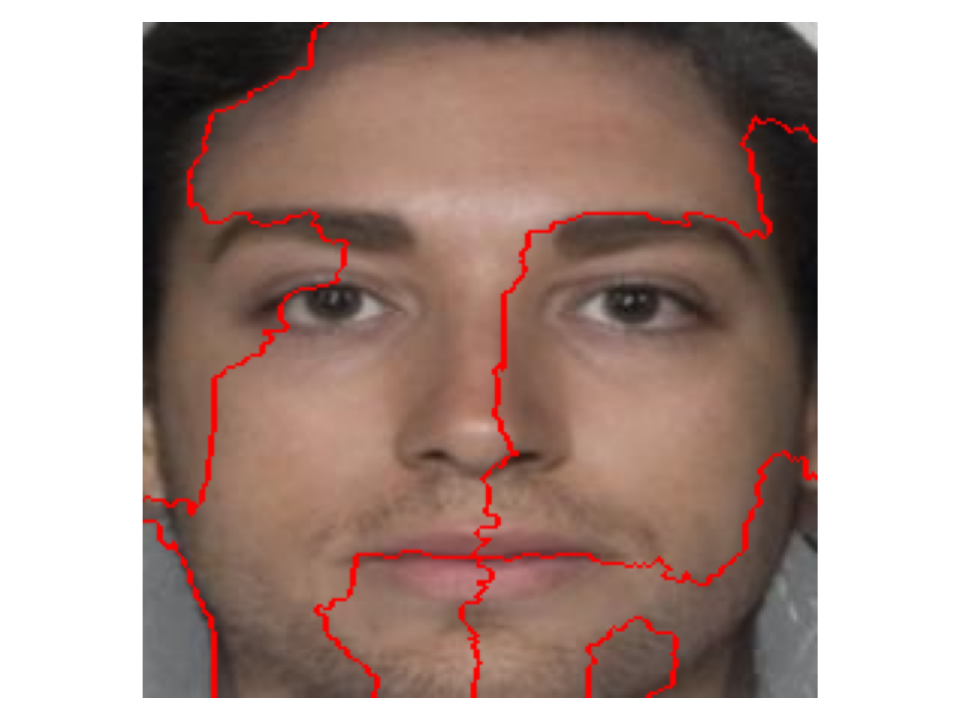}
        \caption{}
        \label{fig:protocol7}
    \end{subfigure}
    \hspace{0.005\textwidth}
    \begin{subfigure}[t]{0.48\textwidth}
        \centering
        \includegraphics[width=\textwidth]{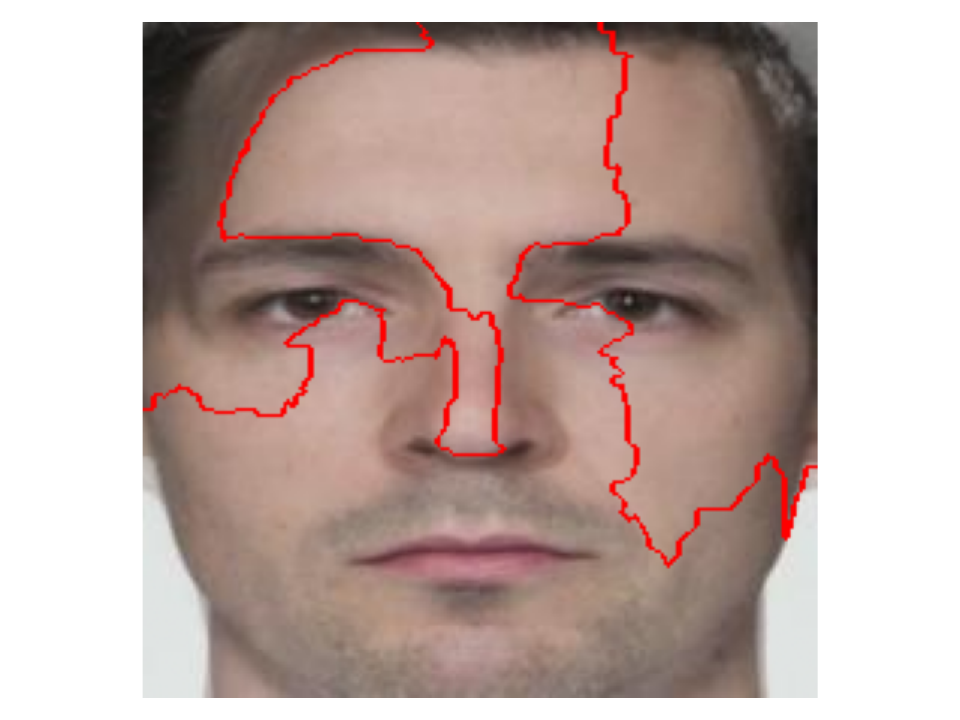}
        \caption{}
        \label{fig:protocol8}
    \end{subfigure}
    \end{minipage}
    }
    \caption{LIME based activation maps: (a) and (b) represent correctly classified bona fide images; (c) and (d) represent misclassified bona fide images; (e) and (f) represent correctly classified morph images; (g) and (h) represent misclassified morph images.}
    \label{fig:lime_maps}
\end{figure}

Figure~\ref{fig:lime_maps} presents LIME-based activation maps for a variety of inputs, illustrating how the model focuses on different facial regions while making its predictions. Subfigures (a) and (b) show correctly classified bona fide images, where the model primarily attends to discriminative facial features such as the periocular region and the mouth, which are often preserved in genuine images. Subfigures (c) and (d) depict misclassified bona fide samples; in these cases, the model appears to be influenced by non-discriminative regions, potentially leading to false positives.

In contrast, subfigures (e) and (f) represent correctly classified morphs, where the model’s attention aligns with typical morphing artifacts such as blending inconsistencies or edge artifacts, particularly around the nose bridge and cheek contours. However, subfigures (g) and (h) display misclassified morph images, where the model fails to detect localized artifacts and instead attends to non-relevant regions, highlighting its limitations in certain morph types or in the presence of subtle manipulations.

These findings offer valuable insights into the model’s reasoning process and help identify both its strengths and vulnerabilities. Furthermore, the use of LIME allows us to assess the reliability of predictions and guide future improvements in model robustness and interpretability.

\section{Conclusion}
\label{sec:Concl}
This paper presents a novel Single-Image Morphing Attack Detection (S-MAD) method using a teacher-student framework, where a ViT-based student model is refined by an EfficientNetV2 teacher model. By incorporating Low-Rank Adaptation (LoRA) for fine-tuning, our approach reduces computational costs while maintaining high detection accuracy. The student model effectively leverages the teacher’s knowledge, enhancing its ability to detect morphing attacks. Evaluation results show that our method outperforms six state-of-the-art S-MAD techniques in both accuracy and efficiency. This demonstrates the potential of knowledge distillation to improve generalization across different morphing techniques while ensuring computational feasibility. Our approach provides a robust and scalable solution for securing Face Recognition Systems against morphing attacks. Future work will explore more dynamic and adaptive uses of the teacher-student framework, such as progressive distillation across multiple teacher models or curriculum learning strategies that gradually transfer knowledge based on the complexity of morphing techniques. Additionally, incorporating task-specific supervision, where different teacher heads guide the student on complementary sub-tasks (e.g., landmark detection, texture analysis), could further enhance detection capabilities. Extending this framework to semi-supervised or unsupervised settings may also enable more robust morph detection with limited labeled data.


%
%
%
\bibliographystyle{splncs04}
\bibliography{FaceMorph_Refereces}
%





\end{document}